**Short-term precipitation prediction using deep learning**

Guoxing Chen[1*] and Wei-Chyung Wang[2]

[1]Department of Atmospheric and Oceanic Sciences and Institute of Atmospheric Sciences, Fudan University, Shanghai, China

[2]Atmospheric Sciences Research Center, University at Albany, State University of New York, New York, USA

*Correspondence to: Guoxing Chen, chenguoxing@fudan.edu.cn

**Accurate weather prediction is essential for many aspects of life, notably the early warning of extreme weather events such as rainstorms. Short-term predictions of these events rely on forecasts from numerical weather models, in which, despite much improvement in the past decades, outstanding issues remain concerning model uncertainties, and increasing demands for computation and storage resources[1,2]. In recent years, the advance of deep learning offers a viable alternative approach[3]. Here, we show that a 3D convolutional neural network using a single frame of meteorology fields as input is capable of predicting the precipitation spatial distribution. The network is developed based on 39-years (1980–2018) data of meteorology and daily precipitation over the contiguous United States. The results bring fundamental advancements in weather prediction. First, the trained network alone outperforms the state-of-the-art weather models in predicting daily total precipitation, and the superiority of the network extends to forecast leads up to 5 days. Second, combining the network predictions with the weather-model forecasts significantly improves the accuracy of model forecasts, especially for heavy-precipitation events. Third, the millisecond-scale inference time of the network facilitates large ensemble predictions for further accuracy improvement. These findings strongly support the use of deep-learning in short-term weather predictions.**

## 1. Introduction

Weather significantly affects human activities. It provides natural resources such as fresh water (for daily consumption, agricultural and industrial uses) and solar radiation and winds (for energy generation). On the other hand, extreme weathers (e.g., hurricanes, rain- and snow-storms) often create dire situations (e.g., floods, landslides and wild fires) causing severe economic losses and casualties. As the global warming continues, the frequency and intensity of extreme weathers are likely to increase in many regions[4]. Thus, accurate weather prediction, particularly the short-term forecast, is indispensable both for management of natural resources and for early warning and mitigating impacts of extreme weather events.

Conventional weather forecast mainly relies on first-principle numerical weather models, which simulate the evolution of atmospheric state using a geophysical fluid dynamic framework coupled with various physical parameterizations. Significant efforts have been invested during the past few decades in observation[5, 6], numerical modeling[7, 8], and data assimilation[9, 10] to improve the model forecasting skill[1, 2]. However, large uncertainties persist for many causes. For example, the precipitation prediction, which involves complex, multi-scale interactions between aerosols, clouds, radiation and large-scale meteorology conditions[11], is still difficult due to inadequate understanding of many physical processes (e.g., ice- and mixed-phase cloud physics) for their representations in models[12]. Moreover, as finer grid resolution and more-realistic physical parameterizations are employed, the demand for computing power and data storage also increases[1], hindering the model operations such as limiting the size of ensemble simulations.

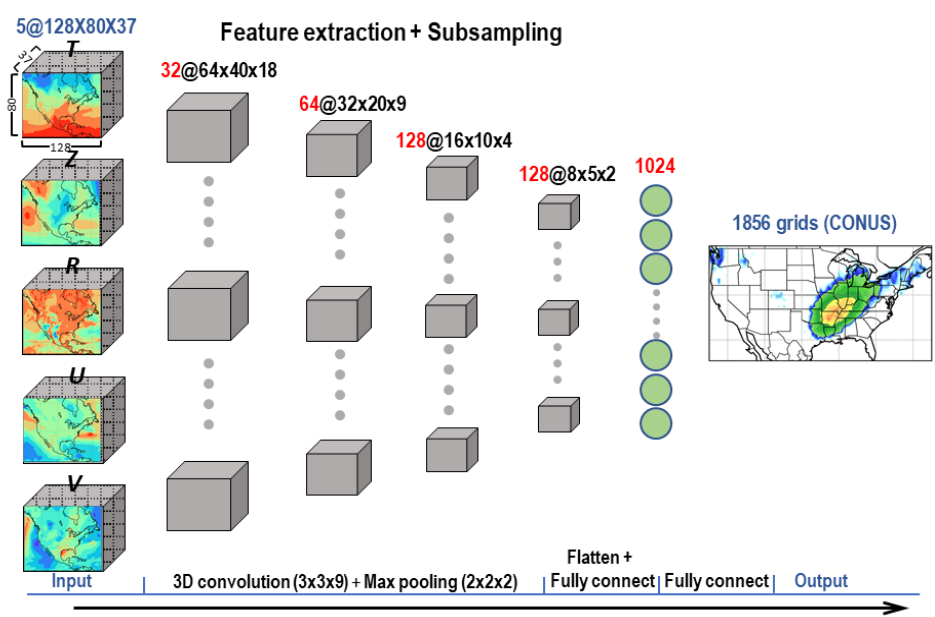

**Fig. 1 | Architecture of the fast precipitation prediction (FPP) neural network.** The network consists of one input layer, four 3D-convolution layers with max-pooling subsampling, one fully-connected layer and one output layer. The input is a single-frame 3D fields of air temperature ($T$), geopotential height ($Z$), relative humidity ($R$), and horizontal wind speeds ($U$ and $V$), and the output is the next-day total precipitation over the contiguous United States (CONUS). The database is 39-years (1980–2018) data of ERA-Interim meteorology and CPC daily precipitation, with 5 years (1997, 2002, 2007, 2012 and 2017) for validation, and 5 years (1998, 2003, 2008, 2013 and 2018) for testing, and the rest 29 years for training. The network is implemented using the Pytorch API and trained on NVIDIA GeForce RTX 3090 graphics cards.

In recent years, the deep-learning methods, which can automatically extract spatiotemporal features in data and reveal physical associations[13], become a useful tool for weather and climate modeling[3, 14]. Its applications range from improving physical processes of radiation[15], convection[16–18] and boundary-layer turbulence[19] to postprocessing model output[20–22], and to building independent prediction models[23–30]. All demonstrate encouraging outcomes. For example, for prediction models, precipitation seasonal forecast[31] and nowcast[32–36] based on deep-learning methods often match and sometimes outperform conventional model forecasts.

This study demonstrates a successful application of deep learning in short-term precipitation prediction. Specifically, we develop a fast precipitation prediction (FPP) neural network for predicting daily total precipitation over the contiguous United States (CONUS). As shown in Fig. 1, the network takes a single frame of 3D (i.e. longitude, latitude and altitude) meteorology analysis fields (temperature, geopotential height, relative humidity, zonal and meridional wind speeds) as input, which is similar to the conventional weather forecast models. Including the full set of meteorology variables ensures that relevant dynamic or thermodynamic factors pertinent to precipitation can be derived from the input. A series of 3D convolutions follow to automatically extract horizontal (e.g., divergence/convergence) and vertical (e.g., instability, convective available potential energy and total water path) features to yield the total precipitation of the following 24-hour period as output. In this approach, the network learns the physical connections between meteorology fields and precipitation, and thus can yield a realistic prediction. Note that the input domain covers the CONUS and surrounding regions, thus carrying enough upstream/downstream information for deriving the lateral-boundary forcing.

We used 39-years data (1980–2018) of ERA-Interim (ERA-I) meteorology[37] and NOAA CPC (Climate Prediction Center) daily precipitation[38–40], in which 29-years were used for training, 5-years for validation, and 5-years for testing the neural network. Detailed descriptions of adopted data and the network architecture are given in the Materials and Methods. Below, based on the testing results, we summarize the key findings about the network skill mainly using the metric of root-mean square error (RMSE) between predictions and CPC observations; here a smaller RMSE indicates a better prediction.

## 2. Results

The RMSE values are calculated for two types of prediction products, the 'raw' prediction (RP) and the postprocessed predictions. The latter consist of: a 'tuned' prediction (TP) with improved skill for heavy precipitation; and a 'weighted' prediction (WP) by combining TP and numerical-model forecasts for further skill improvement (see Method for postprocessing algorithms). For comparisons, we choose three state-of-the-art conventional model forecasts: ERA-Interim 12-hour and 24-hour forecasts (see Method for detailed descriptions), and the MERRA2 forecast[41,42], referred as E12, E24 and M2, respectively. Note that E12 utilizes more observations than real-time forecasts for the 24-hour accumulated precipitation and is not practical for operational forecast; thus in this study, E12 is only considered as the benchmark for assessing other model forecasts and network predictions. In the following, we discuss first the general performance of the FPP network in predicting daily total precipitation over the CONUS, and next focus on the network predictive skill for heavy-precipitation events.

## 2.1 General performance

Figure 2 compares RMSEs of the daily-precipitation predictions from the FPP network and numerical models over the 5-years testing data. Clearly, the FPP raw prediction RP, having the domain-mean RMSE of 4.64, outperforms E24 and M2, whose domain-mean RMSEs are 4.78 and 5.14, respectively. This indicates that the network captures physics that has not been captured in the previous formulations of the first-principles numerical models. The spatial distribution of RP RMSE exhibits quite similar patterns as those of model forecasts, showing larger values over the southeast and along the west coast, and smaller values over the western mountain regions. These patterns resemble the spatial distribution of CPC mean daily-precipitation (depicted by contours in Fig. 2a). This suggests that the large RMSE in the southern regions especially along the coast of Gulf is mainly associated with biases in heavy-precipitation events.

We have also examined RP performance in different regions and seasons. In general, RP outperforms E24 and M2 in the Midwest and South, but has larger biases in the Northeast (see Supplementary Fig. 1). The RMSE over Northeast is relatively larger in autumn and winter caused by the scarcity of snow and mixed-phased precipitation in the training data (see Supplementary Fig. 2). In contrast, the RP skill over Southern region shows much smaller seasonal variations.

To gain more insights into RP, we examine the RMSE at different precipitation intensity levels. As shown in Fig. 2g, RP is very good at predicting light precipitation (1–10 mm day$^{-1}$), but has larger biases at moderate (10–25 mm day$^{-1}$) and heavy (> 25 mm day$^{-1}$) precipitation caused mainly by under-prediction of the heavy-precipitation intensity (see Supplementary Fig. 3). These features can be attributed to the skewed occurrence-frequency distribution of different precipitation-intensity events in the training data, ~ 21% in light precipitation and < 7% in the moderate and heavy precipitation, making the trained network weighed more towards weaker precipitation.

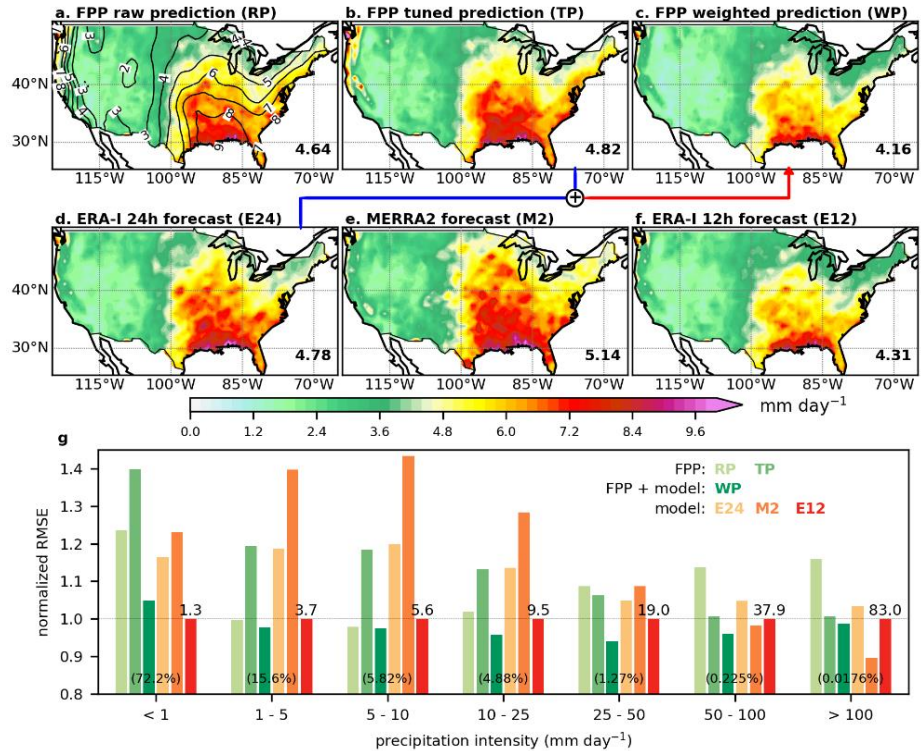

**Fig. 2 | Comparisons of root-mean square error (RMSE) between FPP network predictions (RP, TP, and WP) and the state-of-the-art numerical model forecasts (E24, M2 and E12) over the 5-years testing data.** (a–f) Spatial distribution, with the domain means given at lower-right corners; (g) RMSE at different precipitation intensity, normalized by the respective E12 RMSE, whose values are listed on the 1.0 horizonal line. Contours in (a) are the daily precipitation intensity (mm day$^{-1}$) averaged over 5 testing years; and percentage numbers in (g) are the occurrence frequency of the corresponding intensity level. RP outperforms E24 and M2; TP, better predicting heavy precipitation than RP, is superior to M2 for precipitation intensity of 1–50 mm day$^{-1}$; and WP, the weighting average (indicated by the ⊙ symbol) of E24 and TP, beats E12 for most precipitation intensity levels, indicating that the FPP results can significantly improve the 24-hour model forecast.

Because of the high stakes of heavy-precipitation events (e.g., rainstorms), we postprocessed RP via tuning to yield TP to alleviate the network underestimation of heavy precipitation. Although the domain-mean RMSE increases to 4.82 for TP (Fig. 2b) due to increased RMSE for light precipitation (Fig. 2g), the predictive skill for heavy precipitation is much improved, matching or even beating E24 and M2. Further processing TP using the weights from model forecast E24, the WP yields an RMSE of 4.16 (Fig. 2c), which is even smaller than that of E12 (4.31). In other words, the network prediction can improve the accuracy of 24-hour forecast to the level of 12-hour forecast, showing improvement in most categories of precipitation intensity (Fig. 2g) and in most subregions (see Supplementary Figs. 1, 2). In addition, we found that the combination of TP and E24 shows smaller domain-mean RMSE than the combination of M2 and E24 at any averaging weight (see Supplementary Fig. 4). It indicates that the FPP network prediction can provide more valuable information than a conventional-model forecast within the context of multi-model predictions.

## 2.2 Predictive skill for heavy-precipitation events

To further illustrate the network predictive capability, we identified extremely-heavy precipitation events in the 5-years testing data and examined the network performance on individual rainstorm events. These events, 16 in total, are unevenly distributed in the five years and all have precipitation exceeding 25 mm day$^{-1}$ over one twelfth of the CONUS area. Comparisons of the RMSE and pattern correlation coefficient (PCC; a larger PCC indicates a better prediction) between the FPP network predictions and model forecasts, shown in Supplementary Table 1, reveal that the best predictions are: WP in 9 events, RP in 2, TP in 1, M2 in 1, and E12 in 3. The skill of WP is very impressive, outperforming E24 in 15 events and E12 in more than half of the events. This affirms the significant values of the network predictions to supplementing conventional-model forecasts.

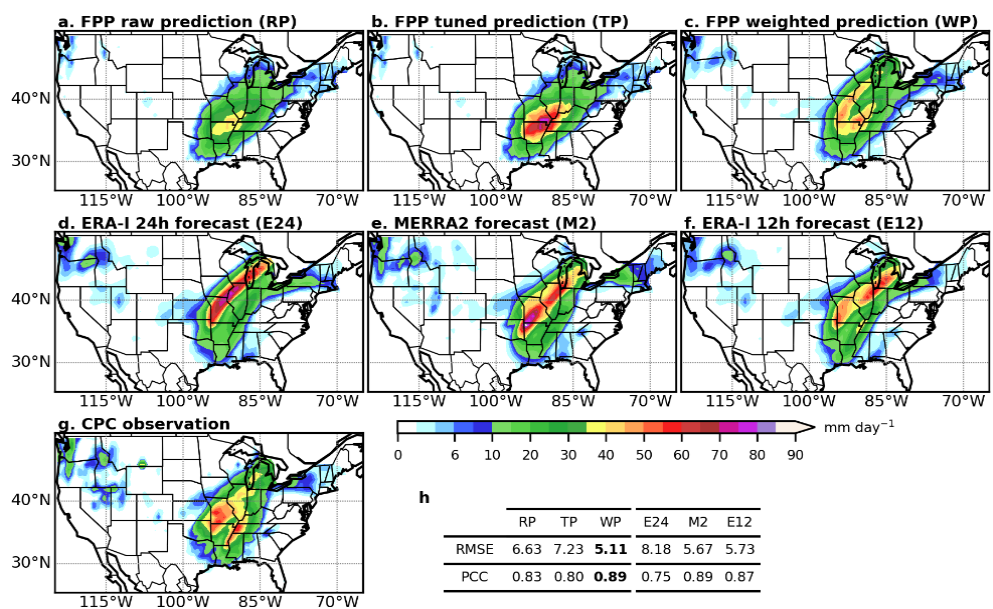

**Fig. 3 | Precipitation distributions from FPP network predictions (a–c) and numerical weather forecasts (d–f) for the 2013/01/30 heavy precipitation event, compared with the CPC observation (g).** The corresponding domain-mean RMSE and pattern correlation coefficient (PCC) are listed in (h). Because of the different characteristics of the predicted distributions from FPP network and conventional models, the combined prediction i.e. WP is the closest to the CPC observation, outperforming E12 with both smaller RMSE and larger PCC.

Figure 3 examines the predicted spatial distribution of the 30 January 2013 heavy-precipitation event. In the CPC observation, the precipitation clearly has two peaks that are located at the lower-right Midwest and the upper-central South, respectively. The FPP results, RP and TP, miss the Midwest peak, capture the South peak and have relatively larger area with moderate precipitation. In contrast, the three model forecasts overestimate the Midwest peak, underestimate the South peak and predict smaller areas with moderate precipitation. These indicate that the FPP network, using almost identical input as the numerical weather models, can provide remarkably good but different predictions due to the observation-informed learning process. Consequently, taking advantage of the large inter-model spread, combining TP and E24 (Fig. 3c) not only significantly improves E24 (Fig. 3d), but also beats E12 (Fig. 3f) with both smaller RMSE and larger PCC (Fig. 3h).

## 3. Conclusion and discussion

We have developed an FPP neural network that is capable of outperforming the state-of-the-art weather forecast models in predicting daily total precipitation (Fig. 2), in particular that of heavy-precipitation events (Fig. 3), over the contiguous United States. Since the inference time of the neural network is in milliseconds, which is several-orders of magnitudes faster than the simulation of numerical weather models, the network can be used to conduct large ensemble predictions. We have evaluated the ensemble predictions of 36 neural networks that differ from one another in input channels or network architectures, and found that the ensemble results are superior to the single-network results TP as well as the conventional-model forecast E24 at most precipitation intensity levels (see Supplementary Fig. 5).

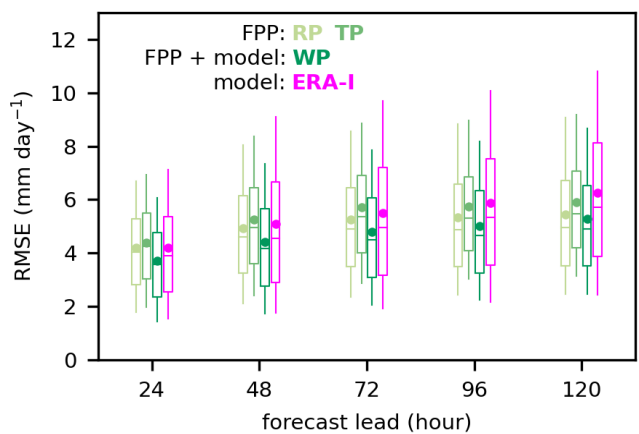

**Fig. 4 | Domain-mean RMSE of FPP network predictions (RP, TP, and WP) for the 5-years testing data at several forecast leads compared with that of the corresponding ERA-Interim (ERA-I) forecast.** The box, whiskers, line and dot show the interquartile range, the 10th–90th percentile range, median and mean of daily domain-mean RMSE, respectively. RP and WP outperform the ERA-Interim forecast at all leads, showing higher superiority with the lead increase.

Finally, although the results above are based on next-day total precipitation, the scope of deep learning approach is much broader, applicable to other weather parameters (e.g., temperature and wind), at different regions, and for shorter or longer forecast leads, as long as sufficient data are readily available for training. For example, for the latter, the network performance for up to 5-days leads (shown in Fig. 4) is always better than the corresponding ERA-Interim forecasts, even though we did not make any modifications to the network architecture or input-domain size, where the optimal configurations may be sensitive to the forecast lead. Therefore, it is concluded that the prospects of deep learning in short-term weather prediction are very promising, and that certainly it is worthy of real-time evaluations side-by-side with conventional models in weather forecast operations.

**Method**

This section is arranged as follows: first a data introduction of the ERA-Interim reanalysis, the CPC observed precipitation and the MERRA2 precipitation forecast; second a description of the neural-network architecture and data usage, and finally the algorithms for postprocessing neural-network prediction and calculating evaluation metrics.

**Data**

The ERA-Interim reanalysis is an estimate for global atmospheric state made by the European Center for Medium-Range Weather Forecasts[37] (available at https://rda.ucar.edu/datasets/ds627.0/). It is produced using a sequential data assimilation scheme that is initialized twice per day at 00Z and 12Z, respectively. At each initialization, available observations of atmospheric state are combined with the forecasted atmospheric state from the prior cycle using the 4D-Var assimilation method to best estimate global atmospheric state, and a short-range model forecast is started to provide state estimates for the next initialization. Each forecast predicts surface precipitation for up to 240 hours. The data has a horizontal resolution of $0.7° \times 0.7°$ (~ 79 km), and 37 vertical levels unevenly extending from 1000 hPa to 1 hPa. It is available for the period of January 1979 to August 2019.

The CPC daily precipitation data over the contiguous United States (CONUS) is a unified gauge-based analysis made by NOAA Climate Prediction Center[38–40] (available at https://psl.noaa.gov/data/gridded/data.unified.daily.conus.html). Values are the accumulated precipitation from 12Z of the day before to 12Z of the day. The data has a horizontal resolution of $0.25° \times 0.25°$, and is available for the period from 1948 to the near present.

MERRA2 is a reanalysis made by the NASA Global Modeling and Assimilation Office using the GEOS-5.12.4 system[41] (available at https://daac.gsfc.nasa.gov/). There are two kinds of precipitation estimates in the reanalysis: the precipitation generated by the atmospheric general circulation model, and the precipitation corrected by merging the observed precipitation[42]. In this study, only the first i.e. the uncorrected precipitation is used as a precipitation forecast made by conventional weather models. The data has a horizontal resolution of $0.5°$ latitude $\times 0.67°$ longitude, and is available for the period from 1979 to the near present.

**Network architecture and data usage**

The fast precipitation prediction (FPP) network, show in Fig. 1, is composed of one input layer, four 3D-convolution layers (kernel size $3 \times 3 \times 9$) with max-pooling subsampling (kernel size $2 \times 2 \times 2$), one fully-connected (FC) layer and one output layer. All 3D convolutions are followed by the ReLU-activation and dropout (p = 0.1) operations; the FC layer is followed by the dropout operation (p = 0.1); and the output layer is followed the ReLU activation (not shown in figure). The MSE (mean square error) loss function is used in training.

The network input is a single frame of 3D meteorology fields including temperature (*T*), geopotential height (*Z*), relative humidity (*R*) and horizontal wind speeds (*U* and *V*), and the output is daily total precipitation over the CONUS. The five input variables are not fully independent. For example, cold temperature at lower levels tends to be accompanied by higher geopotential heights; and horizontal winds at upper levels usually maintain the geostrophic balance and thus can be calculated with gradient of geopotential height. We have examined the channel sensitivity by training the network with one variable excluded and the network architecture unchanged, and the results suggest that excluding any variable tends to reduce the model performance (details are given in Supplementary Fig. 6).

The network database is 39-years data (1980–2018) of ERA-Interim meteorology at 12Z over 7°–63°N, 140°–50°W (80 × 128 grids) and CPC daily precipitation (regridded to ERA-Interim grids). The data are divided into 3 parts: 5 years (1997, 2002, 2007, 2012 and 2017) for validation, 5 years (1998, 2003, 2008, 2013 and 2018) for test, and the rest 29 years for training. This splitting strategy ensures the training data contain different phases of atmospheric multi-year oscillations (e.g., ENSO) (*14*).

In addition, this study uses the ERA-Interim 12-hour (the sum of 12-hour forecasts from the cycles initialized at 12Z and the following 00Z), 24-hour (the 24-hour forecast from the cycle initialized at 12Z) forecasts and the MERRA2 uncorrected forecast for the surface precipitation as references for assessing the performance of the FPP network. Clearly, the ERA-Interim 12-hour forecast utilizes twice observations via initialization than the ERA-Interim 24-hour forecast, while the neural-network prediction utilizes almost identical inputs as the ERA-Interim 24-hour forecast except that the latter further utilizes surface variables such as surface temperature and moisture.

**Postprocessing algorithms**
Because the occurrence frequency of no precipitation and light precipitation is much higher than that of heavy precipitation both spatially and temporally, the trained network is inclined to underestimate the intensity of heavy precipitation, which is more concerned by the public. Therefore, we tune the raw network prediction of precipitation with a simple analytic scheme shown below to enhance the network performance over heavy-precipitation events:

$$TP_j[i] = \left[ 1 + (A-1) \frac{RP_j[i]^3}{\max(RP_j)^3} \right] \cdot RP_j[i],$$

where *A* is the augmentation factor determined on the validation set, representing the differences of the daily maximum and daily mean between the precipitation prediction and the observed value, which is defined as

$$A = 0.5\left(\frac{\langle \max(OB_j) \rangle}{\langle \max(RP_j) \rangle} + \frac{\langle \mathrm{mean}(OB_j) \rangle}{\langle \mathrm{mean}(RP_j) \rangle}\right).$$

In the rest symbols, *i* denotes the longitude-latitude grid index; *j* the date index; $RP_j[i]$, $TP_j[i]$ and $OB_j[i]$ the raw network prediction (i.e. without tuning), tuned network prediction, and observation of precipitation at the $i^{th}$ grid on the $j^{th}$ day, respectively; $\max(X_j)$ the maximum value on the $j^{th}$ day; $\mathrm{mean}(X_j)$ the mean value on the $j^{th}$ day; $\langle X \rangle$ the average over the 5 years in the validation data set (i.e. index *j*).

In addition, the tuned network prediction is combined with the ERA-Interim 24-hour forecast to assess whether and how much the network prediction can improve the prediction of conventional weather models. The combination is realized via weighting average

$$WP_j = 0.5 TP_j + 0.5 E24_j.$$

where $WP_j$ is the weight-averaged prediction, $E24_j$ is the ERA-Interim 24-hour forecast, and the weighting factor (0.5) is determined by scanning over the validation dataset.

**Root-mean square error and pattern correlation coefficient**
The root-mean square error (RMSE) is calculated as

$$RMSE = \sqrt{\frac{\sum_{j=1}^{M}\sum_{i=1}^{N}(P_j[i]-OB_j[i])^2}{M \times N}},$$

and the pattern correlation coefficient (PCC) is calculated as

$$PCC_j = \frac{\sum_{i=1}^{N}\left(P_j[i]-\mathrm{mean}(P_j)\right)\left(OB_j[i]-\mathrm{mean}(OB_j)\right)}{\sqrt{\sum_{i=1}^{N}\left(P_j[i]-\mathrm{mean}(P_j)\right)^2}\sqrt{\sum_{i=1}^{N}\left(OB_j[i]-\mathrm{mean}(OB_j)\right)^2}}.$$

Therein, $P_j[i]$ indicates the network/model prediction for the $i^{th}$ grid on the $j^{th}$ day, *M* the total number of days in the test data, and *N* the total number of grids over the CONUS.

**Supplementary information**

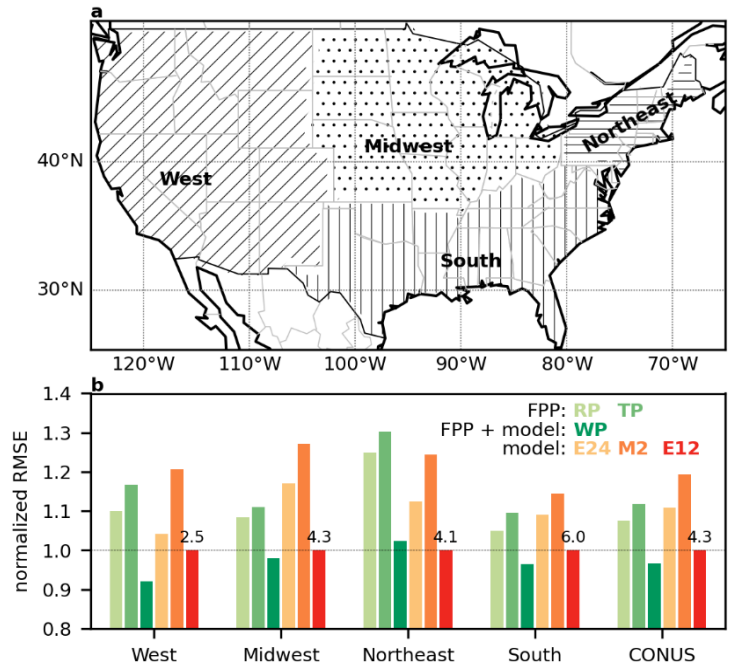

**Supplementary Fig. 1 | Root-mean square error (RMSE) of precipitation predictions over four subregions in the contiguous United States (CONUS).** (a) Definition of the four subregions i.e. West, Midwest, Northeast and South (shown by hatching); (b) RMSE over the subregions and the whole CONUS, normalized with RMSE of the respective ERA-Interim 12-hour forecast (E12), whose values (mm day$^{-1}$) are given at the 1.0 horizontal line. The network performance is relatively better (i.e. the normalized RMSE is smaller) over Midwest and South than over West and Northeast.

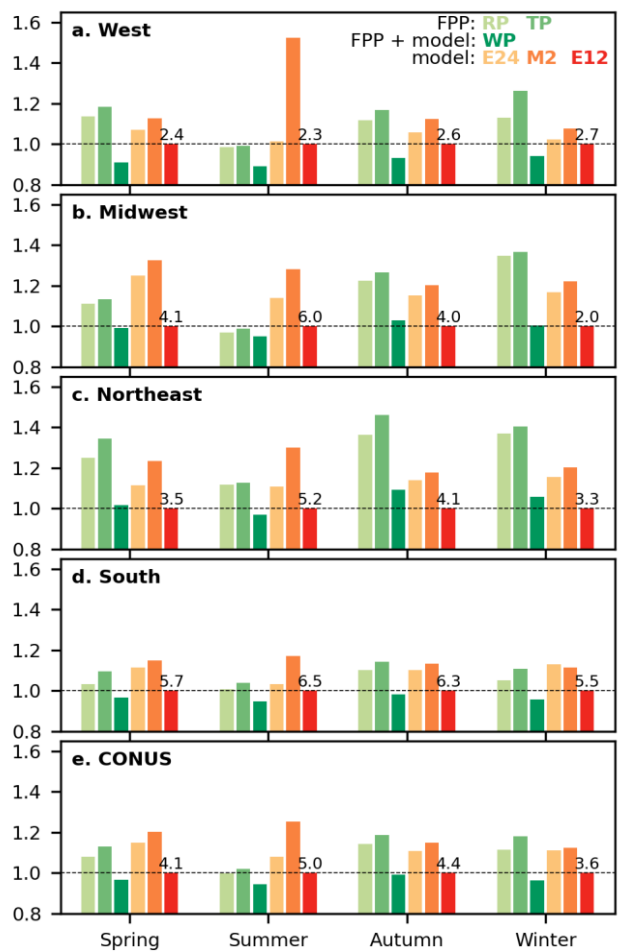

**Supplementary Fig. 2 | Seasonal variation of RMSE over different subregions and the whole CONUS, normalized with the respective E12 RMSE, whose values (mm day$^{-1}$) are given at the 1.0 horizontal line.** The seasons of spring, summer, autumn, winter refer to March-April-May, June-July-August, September-October-November, and December-January-February, respectively. Over Northeast, the normalized RMSE is larger in autumn and winter seasons than in spring and summer seasons. It is thus inferred that the large biases over Northeast could be mainly associated with ice- and mixed-phase precipitation events, which frequently occur in the cool season. In contrast, the South region seldom has these kinds of precipitation and shows much smaller seasonal variations in the normalized RMSE.

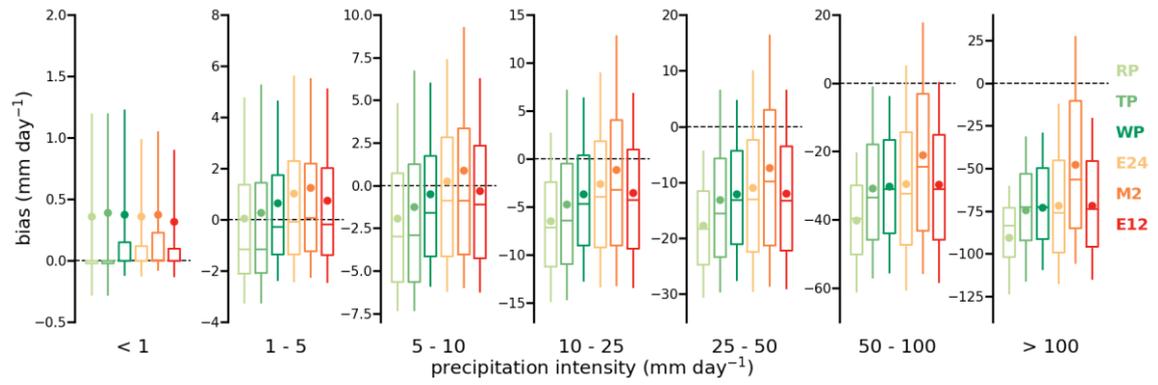

**Supplementary Fig. 3 | The statistics of prediction bias for precipitation of different intensity levels (mm day$^{-1}$).** The box, whiskers, line and dot show the interquartile range, the 10th–90th percentile range, median and mean, respectively.

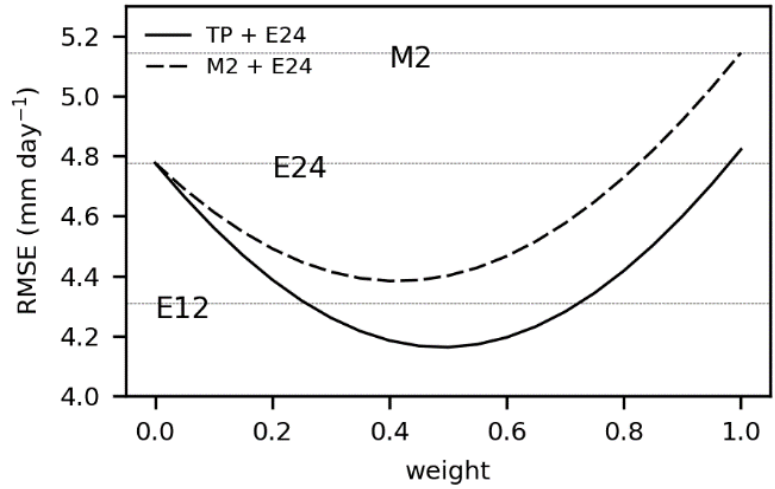

**Supplementary Fig. 4 | RMSE of the TP+E24 combination and the M2+E24 combination as a function of averaging weight.** The combination of M2 and E24 can have smaller RMSE than E24 at certain weights, but it is always inferior to the combination of TP and E24 and never beats ERA-Interim 12-hour forecast (E12). This indicates that the network prediction can add more valuable information to the ensemble than a conventional-model forecast within the context of multi-model predictions.

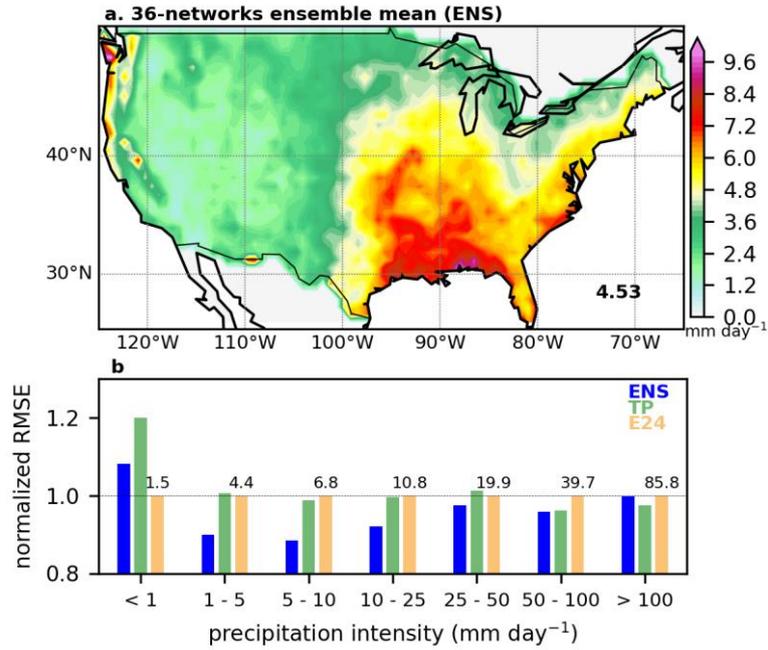

**Supplementary Fig. 5 | RMSE of daily total precipitation from the ensemble mean of 36 neural networks that differ from one another in input channels or network architectures (vs. the CPC observation).** (a) RMSE spatial distribution, with the domain-mean RMSE given at the lower-right corner; (b) RMSE for grid precipitation of different intensity categories, normalized with the respective E24 RMSE, whose values are given at the 1.0 horizontal line. Results from all networks were tuned using identical algorithms described in Materials and Methods before calculating the ensemble mean. The ensemble-mean results (ENS) have even smaller RMSE (4.53) than the conventional model-forecast E24 (4.78) and the single-network prediction TP (4.82). Particularly, the ENS alone can beat E24 at most precipitation intensity categories.

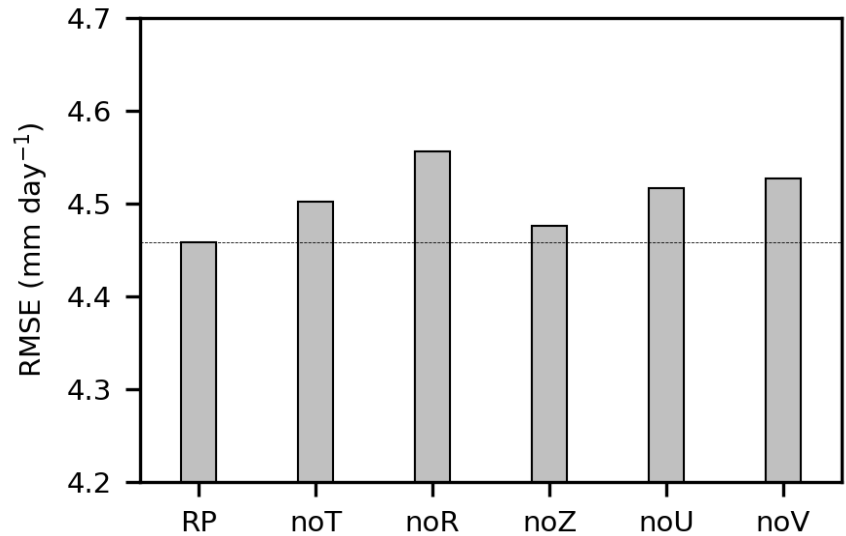

**Supplementary Fig. 6 | Channel sensitivity of the precipitation prediction, estimated based on the validation dataset.** RP utilizes all five channels while the other five predictions (i.e. noT, noR, noZ, noU and noV) exclude one channel, respectively. Excluding any channel weakens the network performance i.e. causes larger RMSE. Therein, excluding geopotential height (noZ) causes the smallest performance decline, whereas excluding relative humidity (noR) or wind speeds (noU, noV) causes much larger declines. This quite makes sense, as the level of relative humidity determines water vapor abundance that is crucial for forming precipitation, while horizontal winds not only affect water vapor transport but also can indicate the migration of large-scale meteorology conditions. Note that this is only channel sensitivity of the current network architecture rather than a general conclusion that may be applied to other architectures.

**Supplementary Table 1 | RMSE (mm day$^{-1}$) and pattern correlation coefficient (PCC) of precipitation distribution of 16 extremely-heavy precipitation events in the 5-years test predictions.** An extremely-heavy precipitation event is defined as a day with more than 150 grids (about 1/12 of the CONUS area) having precipitation exceeding 25 mm day$^{-1}$. There are 16 events in total for the 5-years test period. The best prediction (the one with smallest RMSE and largest PCC; highlighted with underlined bold fonts) is WP in 9 events, RP in 2, TP in 1, M2 in 1, and E12 in 3. WP is superior to E24 in 15 events.

| Date | FPP predictions | | | Conventional forecasts | | |
|---|---|---|---|---|---|---|
| | RP | TP | WP | E24 | M2 | E12 |
| | RMSE/PCC | RMSE/PCC | RMSE/PCC | RMSE/PCC | RMSE/PCC | RMSE/PCC |
| 1998-01-08 | 8.47/0.81 | 8.29/0.80 | **<u>6.70</u>**/**<u>0.87</u>** | 9.59/0.78 | 8.43/0.82 | 8.39/0.81 |
| 1998-03-08 | 10.09/0.78 | 10.39/0.80 | 9.86/0.80 | 12.82/0.65 | **<u>9.77</u>**/**<u>0.82</u>** | 11.86/0.68 |
| 1998-03-09 | **<u>7.19</u>**/**<u>0.84</u>** | 7.18/0.83 | 7.88/0.79 | 10.77/0.65 | 9.67/0.76 | 10.49/0.65 |
| 1998-10-05 | 10.57/0.65 | 10.76/0.64 | 10.26/0.68 | 11.51/0.63 | 11.75/0.62 | **<u>9.46</u>**/**<u>0.73</u>** |
| 2003-11-18 | **<u>8.17</u>**/**<u>0.82</u>** | 8.60/0.80 | 9.03/0.77 | 11.06/0.67 | 8.32/0.81 | 8.80/0.78 |
| 2003-11-19 | 6.88/0.83 | 7.51/0.80 | **<u>6.02</u>**/**<u>0.87</u>** | 7.09/0.84 | 7.97/0.83 | 6.72/0.84 |
| 2008-09-14 | 12.96/0.64 | 12.98/0.65 | **<u>7.91</u>**/**<u>0.89</u>** | 8.50/0.87 | 8.20/0.88 | 8.27/0.88 |
| 2008-12-12 | 9.16/0.80 | 9.55/0.76 | 6.26/0.91 | 5.84/0.92 | 4.87/0.94 | **<u>4.73</u>**/**<u>0.95</u>** |
| 2013-01-30 | 6.63/0.83 | 7.23/0.80 | **<u>5.11</u>**/**<u>0.89</u>** | 8.18/0.75 | 5.67/0.89 | 5.73/0.87 |
| 2013-09-21 | 10.85/0.80 | **<u>10.37</u>**/**<u>0.81</u>** | 11.10/0.78 | 15.43/0.63 | 12.83/0.75 | 11.29/0.77 |
| 2013-11-27 | 6.10/0.92 | 6.14/0.90 | **<u>5.05</u>**/**<u>0.94</u>** | 6.48/0.92 | 7.06/0.89 | 5.35/0.92 |
| 2013-12-22 | 9.21/0.79 | 10.50/0.75 | **<u>6.56</u>**/**<u>0.89</u>** | 9.72/0.80 | 8.09/0.84 | 8.51/0.83 |
| 2018-04-16 | 7.29/0.82 | 7.40/0.80 | **<u>4.99</u>**/**<u>0.91</u>** | 6.46/0.86 | 6.05/0.88 | 5.94/0.88 |
| 2018-10-10 | 8.81/0.66 | 9.20/0.61 | 8.19/0.69 | 9.29/0.66 | 8.92/0.70 | **<u>7.15</u>**/**<u>0.77</u>** |
| 2018-11-01 | 6.90/0.86 | 8.54/0.85 | **<u>6.86</u>**/**<u>0.88</u>** | 8.39/0.81 | 7.44/0.86 | 7.34/0.85 |
| 2018-12-27 | 6.92/0.84 | 6.80/0.85 | **<u>4.78</u>**/**<u>0.93</u>** | 5.55/0.90 | 7.38/0.85 | 5.72/0.90 |